\documentclass[runningheads]{llncs}

\usepackage[T1]{fontenc}

\usepackage{hyperref}
\usepackage{algorithm}%
\usepackage{algorithmicx}%
\usepackage{algpseudocode}%
\usepackage{placeins}
\usepackage{graphicx}
\usepackage{amsmath}
\usepackage{booktabs}
\usepackage{xcolor}

\begin{document}
\title{Landscape-Aware Automated Algorithm Configuration using Multi-output Mixed Regression and Classification}
\titlerunning{Landscape-Aware AAC using Mixed Regression and Classification}

\author{\underline{Fu Xing Long}\inst{1,2}\orcidID{0000-0003-4550-5777} \and
Moritz Frenzel\inst{3}\orcidID{0000-0002-4025-8773} \and
Peter Krause\inst{4}\orcidID{0000-0001-8302-0100} \and
Markus Gitterle\inst{5}\orcidID{0000-0001-8760-1682} \and
Thomas B\"{a}ck\inst{1}\orcidID{0000-0001-6768-1478} \and
Niki van Stein\inst{1}\orcidID{0000-0002-0013-7969}
}
\authorrunning{F.X. Long et al.}

\institute{LIACS, Leiden University, Niels Bohrweg 1, 2333 Leiden, Netherlands\\
\email{\{f.x.long,t.h.w.baeck,n.van.stein\}@liacs.leidenuniv.nl} \and
BMW Group, Knorrstra{\ss}e 147, 80788 Munich, Germany\\
\email{fu-xing.long@bmw.de} \and
Altair Engineering GmbH, Calwer Straße 7, 71034 B\"{o}blingen, Germany\\
\email{mfrenzel@altair.com} \and
divis intelligent solutions GmbH, Joseph-von-Fraunhofer-Stra{\ss}e 20, 44227 Dortmund, Germany\\
\email{krause@divis-gmbh.de} \and
Munich University of Applied Sciences, Dachauer Stra{\ss}e 98b, 80335 Munich, Germany\\
\email{markus.gitterle@hm.edu}}
\maketitle            
\begin{abstract}

In landscape-aware algorithm selection problem, the effectiveness of feature-based predictive models strongly depends on the representativeness of training data for practical applications.
In this work, we investigate the potential of randomly generated functions (RGF) for the model training, which cover a much more diverse set of optimization problem classes compared to the widely-used black-box optimization benchmarking (BBOB) suite.
Correspondingly, we focus on automated algorithm configuration (AAC), that is, selecting the best suited algorithm and fine-tuning its hyperparameters based on the landscape features of problem instances.
Precisely, we analyze the performance of dense neural network (NN) models in handling the multi-output mixed regression and classification tasks using different training data sets, such as RGF and many-affine BBOB (MA-BBOB) functions.
Based on our results on the BBOB functions in $5d$ and $20d$, near optimal configurations can be identified using the proposed approach, which can most of the time outperform the off-the-shelf default configuration considered by practitioners with limited knowledge about AAC.
Furthermore, the predicted configurations are competitive against the single best solver in many cases.
Overall, configurations with better performance can be best identified by using NN models trained on a combination of RGF and MA-BBOB functions.

\keywords{Black-box optimization \and Exploratory landscape analysis \and Multi-output mixed regression and classification \and Dense neural network \and Randomly generated functions.}
\end{abstract}

\section{Introduction}\label{sec:intro}

In landscape-aware algorithm selection problem (ASP)~\cite{asp_munoz2015,vanstein2024explainable}, the performance of optimization algorithms has been linked to the landscape characteristics of black-box optimization (BBO) problems that are quantified using fitness landscape analysis~\cite{review_malan2021}.
By constructing machine learning (ML) models, for instance, the performance of optimization algorithms can be estimated based on the landscape characteristics of problem instances~\cite{ela_kerschke2019}.
In other words, the problem landscape characteristics can be exploited to select well-performing optimization algorithms from an algorithm portfolio prior to the actual optimization runs.
Using a large set of problem instances, and preferably from diverse optimization problem classes, the corresponding landscape characteristics and algorithm performances are utilized for the training of ML models.
Following this, the effectiveness of predictive models is heavily dependent on the representativeness of training data for unseen BBO problems.

Although landscape features are informative in explaining algorithm performances~\cite{ela_simoncini2018}, landscape-aware ASP was mainly investigated on benchmarking problems in previous work, such as the widely-used black-box optimization benchmarking (BBOB) suite~\cite{bbob_hansen2009_noiseless}.
The fact that the BBOB suite is not representative enough for engineering applications, such as crashworthiness optimization~\cite{long2022learning,long2024generating} and control system calibration~\cite{thomaser2022one} in the automotive industry, raises concerns that predictive models trained using only the BBOB suite might generalize poorly to unseen problem classes that are not being covered.
Moreover, for real-world BBO problems that require expensive function evaluations, e.g., time-consuming and/or costly simulation runs, the function evaluation budget can be particularly limited, hindering the generation of a large data set for the model training.
To fill in the gap, we explore an alternative in building pre-trained general purpose models that can generalize well for different applications, including expensive BBO problems, while maintaining an affordable computational effort.
Our ultimate vision is to assist practitioners with little domain knowledge about ASP, e.g., from engineering fields, to automatically identify the best suited algorithm configuration for their applications.

\paragraph{Our contribution:}
In this work, we investigate the potential of tree-based randomly generated functions (RGF) for the training of predictive models, which are much more diverse than the BBOB suite in terms of optimization landscape characteristics.
In this context, we implement a selection process to identify RGF that are appropriate as training data.
Furthermore, we extend our investigations towards landscape-aware automated algorithm configuration (AAC) by combining both algorithm selection and hyperparameter optimization (HPO), that is, finding the best suited algorithm and fine-tuning its hyperparameters.
For the prediction of optimal configurations, we consider dense neural network (NN) models, which can easily handle multi-output mixed regression and classification tasks.
Based on our empirical results, near optimal configurations can be identified using the proposed approach, which can outperform the off-the-shelf default configuration and compete against the single best solver (SBS) for many BBOB functions.
In some cases, NN models can perform better than random forest (RF) models, which are typically considered for landscape-aware ASP. \\

This paper has the following structure:
Related works are introduced in Section~\ref{sec:sota}, followed by the explanations of our methodology in Section~\ref{sec:method} and experimental setup in Section~\ref{sec:experiment}.
Next, results are analyzed and discussed in Section~\ref{sec:result}.
Lastly, conclusions and future works are presented in Section~\ref{sec:conclusion}.

\section{Related Work}\label{sec:sota}

The idea of using RGF for the training of feature-based predictive models has been previously investigated, such as in~\cite{skvorc2022transfer}.
In summary, it was reported that RGF were ineffective for the training of high-quality models in terms of prediction accuracy.
Independently of the previous work, our work differs mainly in the following extensions.
\begin{enumerate}

\item Instead of simply using any RGF, we implement an intermediate step to select RGF that are appropriate for the model training purposes.
We argue that this step is crucial to improve model accuracy, as discussed in Section~\ref{sec:select}. 
In fact, we suspect that this might partly explain the low model accuracy in~\cite{skvorc2022transfer}.

\item We propose to consider NN-based predictive models to handle the multi-output mixed regression and classifications tasks in AAC, which can sometimes perform slightly better than RF models, refer to Section~\ref{sec:model}.

\item Rather than just selecting the best algorithm from a portfolio of limited algorithms, as typically done in ASP, we extend our investigations towards combined algorithm selection and hyperparameter optimization (CASH)~\cite{thornton2013auto}, or we call AAC~\cite{vanstein2024explainable} in this work. 

\end{enumerate}

\subsection{Automated Algorithm Configuration}\label{sec:sota_hpo}

To tackle AAC problems, where the search space can be a mix of continuous, integer, categorical, and conditional variables, various optimization algorithms have been implemented, such as tree-structured Parzen estimator (TPE)~\cite{bergstra2011algorithms} and sequential model-based algorithm configuration (SMAC)~\cite{lindauer2022smac3}.
As a variant of Bayesian optimization~\cite{mockus1982_bo}, TPE utilizes Parzen estimators as surrogate models, which can handle mixed-integer search space and scale well to high dimensionality.
For example, TPE has been previously applied to fine-tune the learning rates of covariance matrix adaptation evolutionary strategy (CMA-ES)~\cite{zhao2018tuning}.

In this work, we focus on fine-tuning the configuration of modular CMA-ES~\cite{denobel2021}, developed based on the original CMA-ES algorithm~\cite{hansen1996adapting,hansen2001completely}.
In short, different variants, such as active learning, mirrored sampling, threshold convergence, and recombination weights, are integrated as modules that can be individually activated or deactivated, allowing a custom instantiation of CMA-ES.
Subsequently, modular CMA-ES offers a convenient examination of the interactions between different modules as well as between modules and hyperparameters, e.g., population size and learning rates.

\subsection{Black-Box Optimization Benchmarking}\label{sec:sota_bbob}

In previous work, landscape-aware ASP was commonly investigated based on BBO benchmarking suites, such as the well-known BBOB suite~\cite{bbob_hansen2009_noiseless} available in the comparing continuous optimizers (COCO) platform~\cite{coco_hansen2021} and iterative optimization heuristics profiler tool (IOHProfiler)~\cite{ioh_doerr2018}.
Altogether, the BBOB suite consists of $24$ single-objective, continuous, and noiseless functions of different optimization landscape complexity, which we refer to this suite as~\emph{the} BBOB.

Principally, the BBOB functions can be scaled up to arbitrary dimensionality and different problem instances can be created through transformations of the search space and objective values, which is controlled by an internal identifier.
Typically, investigations based on the BBOB suite are carried out within the box-constrained search space $[-5,5]^d$, where the global optimum is located within $[-4,4]^d$, where $d$ represents the dimensionality.
Extensive analysis of the BBOB problem instances is available in~\cite{long2023bbob}.

To complement the diversity of the BBOB suite, additional functions can be generated via affine combination of two BBOB functions~\cite{dietrich2022increasing}, which is based on an interpolation between two selected BBOB functions and uses a weighting factor to control the shifting between functions.
This approach was later generalized to affine combinations of many BBOB functions, also known as many-affine BBOB (MA-BBOB) functions, where the affine combination is no longer limited to only two functions~\cite{vermetten2023mabbob,affine_bbob_gecco}.

\subsection{Randomly Generated Functions}\label{sec:sota_rgf}

Apart from the benchmarking suites, a set of functions can be generated using the function generator proposed in~\cite{tian2020recommender}, covering a diverse set of optimization problem classes, as shown in~\cite{ela_skvorc2021}.
By using a set of selection pressures, mathematical operands and operators are randomly selected from a predefined pool to construct tree-structured function expressions, which we call RGF.
In fact, RGF with similar landscape characteristics to automotive crashworthiness optimization problems can be created, which is lacking in the BBOB suite~\cite{long2024generating}.

Nonetheless, the properties of RGF are not known a priori, e.g., the global optimum and optimization complexity, as oppose to the well-studied BBOB suite.
To tackle this problem, an extension has been attempted on this function generator to guide the function generation towards specific optimization complexity using genetic programming~\cite{long2023challenges}, which is beyond the scope of this work.

\subsection{Exploratory Landscape Analysis}\label{sec:sota_ela}

In landscape-aware ASP, exploratory landscape analysis (ELA) is one of the popular approaches employed to numerically quantify the high-level landscape characteristics of continuous optimization problems, such as multi-modality and global structure.
While initially only six fundamental feature classes were proposed in ELA, namely $y$-distribution, level sets, meta-models, local searches, curvature, and convexity~\cite{ela_mersmann2010,ela_mersmann2011}, more feature classes have been progressively proposed to complement them, e.g., dispersion, nearest better clustering (NBC), principal component analysis (PCA), linear models, and information content of fitness sequences (ICoFiS)~\cite{flacco_kerschke2019,ela_kerschke2015,ela_lunacek2006,ela_munoz2015}.

In brief, a design of experiments (DoE) is required for the ELA features computation, consisting of some samples $\mathcal{X} = \{x_1,\cdots,x_n\}$ and objective values $\mathcal{Y} = \{y_1,\cdots,y_n\}$, which are computed using an objective function $f$, i.e., $f \colon \mathcal{X} \rightarrow \mathcal{Y}$, where $x_i \in \bbbr^d$, $y_i \in \bbbr$, and $n$ is the sample size.
Consequently, the effectiveness of ELA features can be dependent on the DoE sample size, dimensionality, and sampling strategy~\cite{ela_renau2020}.
To overcome potential bias of the hand-crafted ELA features in capturing specific landscape characteristics, deep NN-based methods have been explored to characterize BBO problems based on latent space features, e.g., DoE2Vec~\cite{van2022doe2vec}, which we leave for future work.

\section{Methodology}\label{sec:method}

\begin{figure*}[!htbp]
 \centering
 \includegraphics[width=1.\linewidth,trim=0mm 0mm 0mm 0mm,clip]{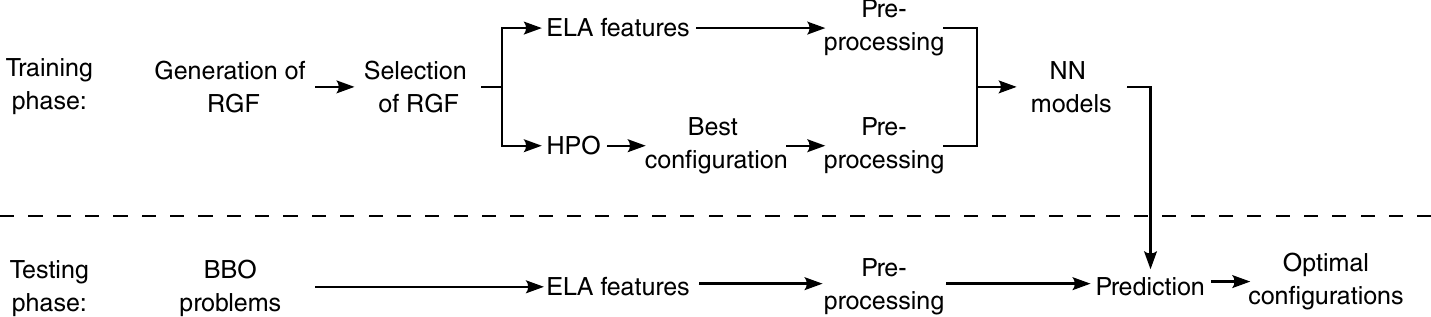}
 \caption{
 An overview of our proposed landscape-aware AAC approach that can identify optimal configurations for BBO problems, consisting of a training and testing phase.
 During the training phase, using a preferably large set of RGF, the respective ELA features and optimal configurations identified through HPO (performed on RGF) are utilized to train NN models.
 The pre-trained models can then be deployed to predict the best suited configuration for unseen BBO problems based on their ELA features in the testing phase.
 }
 \label{fig:pipeline}
\end{figure*}

The workflow of our landscape-aware AAC approach is visualized in Figure~\ref{fig:pipeline}.
In the first step, the landscape characteristics of RGF are captured using ELA and the corresponding best configurations are identified using HPO during the training phase.
Next, the ELA features and configurations are properly pre-processed for the training of NN models.
Eventually, optimal configurations for unseen BBO problems can be predicted based on their ELA features using the trained models.
Our approach is described in detail in the following.

\paragraph{Generation and selection of RGF.}
Firstly, a large set of RGF is generated for the training of NN models, using the function generator implemented in~\cite{long2024generating}.
Before the model training, a pre-selection step is integrated to identify RGF that are appropriate for AAC purposes, refer to Section~\ref{sec:select} for detailed explanations.

\paragraph{Computation of ELA features.}
Secondly, the optimization landscape characteristics of RGF are computed using ELA based on some DoE samples.
To combat inherent bias~\cite{prager2023nullifying}, the objective values are normalized using min-max scaling before the ELA features computation.
Since many of the ELA features are redundant~\cite{ela_renau2019}, highly correlated ELA features based on Pearson's correlation coefficient ($>0.95$) are discarded, using a similar approach as in~\cite{long2024generating}.
To improve the performance of NN models, we ensure that the remaining ELA features are within a comparable scale range via normalization using min-max scaling.

\paragraph{Identifying the best configuration using HPO.}
For each individual RGF, the best performing algorithm configuration found using HPO is considered as the best suited configuration identified for that function.
Similar to the ELA features, the configuration data are pre-processed for the model training, where categorical hyperparameters are one-hot encoded, while continuous hyperparameters are linearly re-scaled to the scale range of $[0,1]$ using Equation~\ref{eq:rescale}.
\begin{equation} \label{eq:rescale}
    z_{new}=\frac{z_{init}-a_{min}}{a_{max}-a_{min}}\cdot \left ( b_{max}-b_{min} \right )+b_{min}\,,
\end{equation}
where $z_{init}$ and $z_{new}$ are the initial and re-scaled values, $a_{max}$ and $a_{min}$ are the lower and upper bound before re-scaling, and $b_{max}$ and $b_{min}$ are the lower and upper bound after re-scaling.
In this work, we focus on finding the best configuration of modular CMA-ES.

\paragraph{Training of NN models.}
For the training of NN models, the pre-processed ELA features are employed as input, while the best configurations identified using HPO as output.
Detailed descriptions of the NN models are included in Section~\ref{sec:model}.

\paragraph{Optimal configurations for BBO problems.}
During the deployment or testing phase, the trained NN models can be used to predict optimal configurations of modular CMA-ES for unseen BBO problems based on their ELA features.
Similar to the training phase, the input ELA features of BBO problems are normalized, while the predicted configurations are inversely transformed back to the original configuration search space.
To avoid invalid configurations, e.g., negative population size, predicted continuous hyperparameters that fall outside the search space will be set to either the lower or upper boundary.

\subsection{Selection of Appropriate RGF}~\label{sec:select}

\begin{figure*}[!htbp]
 \centering
 \includegraphics[width=.33\linewidth,trim=0mm 0mm 0mm 0mm,clip]{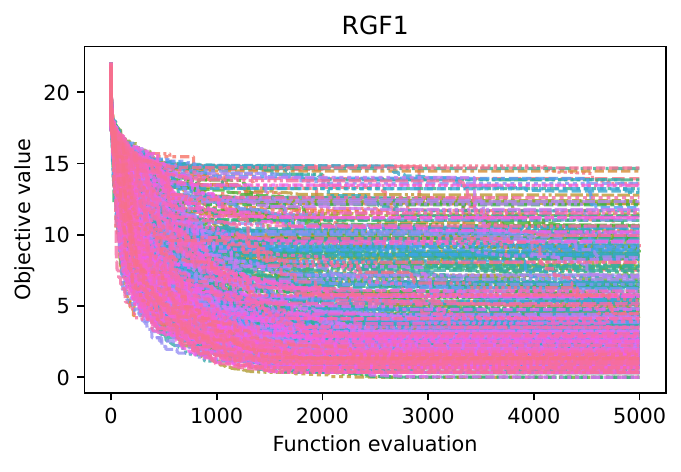}
 \includegraphics[width=.32\linewidth,trim=7mm 0mm 0mm 0mm,clip]{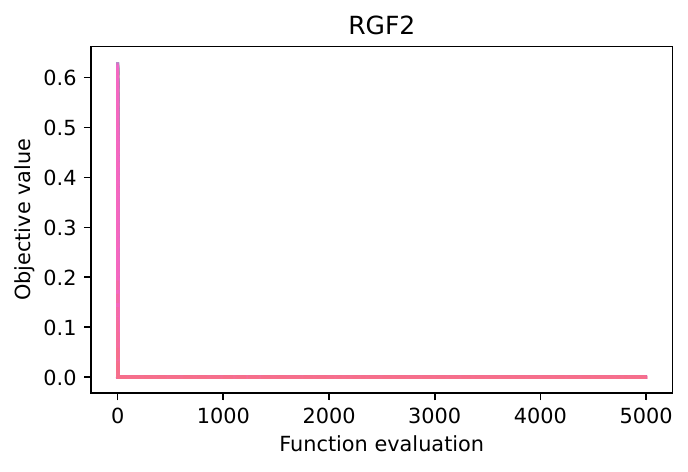}
 \includegraphics[width=.32\linewidth,trim=7mm 0mm 0mm 0mm,clip]{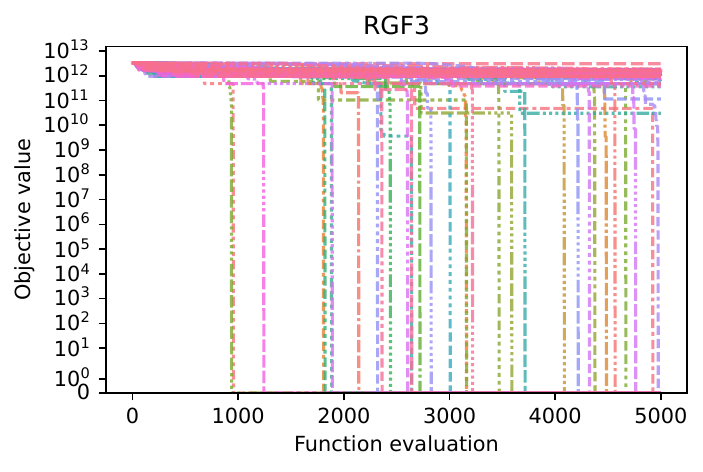}
 \caption{
 The optimization convergence of $500$ configurations evaluated using HPO on three chosen RGF.
 The x-axis shows the number of function evaluations, while the y-axis shows the re-scaled objective values, with $0$ being the best solution found in all runs.
 Each curve represents a configuration run using modular CMA-ES (median over $10$ repetitions).
 (\textit{Left}) Ideal for AAC purposes, where a clear ranking of configurations is possible.
 (\textit{Middle}) Ambiguous ranking of algorithm configurations, where all configurations are equally competitive.
 (\textit{Right}) The global optimum seems to be an outlier that can only be found by a few configurations.
 }
 \label{fig:bad_rgf}
\end{figure*}

Unlike the well understood BBOB functions, the landscape characteristics and global optimum of RGF are not known a priori.
Due to the fact that some RGF are insufficiently discriminative in distinguishing different configurations based on their optimization performances, not all RGF are appropriate for AAC purposes based on our preliminary testing.
Using the HPO results on three chosen RGF in Figure~\ref{fig:bad_rgf} as an example, we consider functions with a similar pattern to `RGF1' as ideal for AAC purposes, where a clear configuration ranking with only a few ties is possible.
More importantly, the best configuration can be easily identified.
On the other hand, functions similar to `RGF2' are considered as inappropriate for AAC, where many, or in extreme situations, all configurations are equally good, leading to an ambiguous ranking.
We suspect that the optimization complexity of such RGF is too low that the choice of configuration does not matter.
Surprisingly, two RGF with a small difference in their ELA features can have the opposite patterns, which raises questions for future research.
To improve the robustness of trained models, functions similar to `RGF3' are additionally neglected, where the global optimum seems to be an extreme outlier and can be found occasionally by a few configurations.

To overcome these problems, the following measures are implemented to identify RGF that are appropriate for AAC purposes.

\begin{enumerate}
\item \textbf{Estimation of global optimum:}
In a brute-force manner, we perform HPO on each RGF, focusing on finding a better solution, i.e., a smaller objective value, and using a similar setup as described in Section~\ref{sec:experiment}.
Eventually, the global optimum $y_{opt}$ is approximated based on the best solution found in all HPO runs $y_{hpo}$ using Equation~\ref{eq:yopt}.
\begin{equation}\label{eq:yopt}
\begin{split}
y_{opt} &=
\begin{cases}
\left \lfloor y_{hpo} \right \rfloor, & \text{ if } 0\leq \left | y_{hpo} \right | < 10 \\
\left \lfloor y_{hpo}/10 \right \rfloor\cdot 10, & \text{ if } 10\leq \left | y_{hpo} \right | < 100 \\ 
\left \lfloor y_{hpo}/10^p \right \rfloor\cdot 10^p, & \text{ otherwise  } \\
\end{cases}
, \\
p &= \left \lfloor \log_{10}\left | y_{hpo} \right | \right \rfloor - 1 , 
\end{split}
\end{equation}
where~$y_{hpo}$ is either rounded to the nearest lower integer for a small~$\left | y_{hpo} \right |$, or rounded based on the nearest lower power of $10$.
Having an estimated global optimum for RGF is essential in our approach to facilitate an evaluation of configuration performance (refer to Section~\ref{sec:metric}) and a comparison between different functions with varying scale ranges.

\item \textbf{RGF appropriate for AAC:}
Using the same HPO results from previous step, all configurations evaluated are ranked according to their performances, where ties are assigned with the same rank.
The ranking ambiguity is evaluated based on the Kendall rank correlation coefficient between the HPO configuration ranking and a strict ranking (without tie).
For a correlation lower than $0.9$, e.g., due to too many ties, such ranking is considered as ambiguous.
Furthermore, we compute the standard score or z-score of the global optimum found to estimate its deviation from the distribution of other solutions.
When the global optimum is $3$ standard deviations away from the distribution mean, it is considered as an extreme outlier.

\item \textbf{Elimination of RGF:}
A RGF is excluded from the training data, if any of the aforementioned conditions is fulfilled.

\end{enumerate}

While additional computational effort is required for the above-mentioned measures in identifying RGF appropriate for AAC purposes, we argue that they are critical in improving the performance of NN models.
Moreover, this process needs to be done only once, since the RGF identified can be re-used in the future for the same BBO problem classes.

\subsection{Multi-output Mixed Regression and Classification}~\label{sec:model}

\begin{figure}[!htbp]
 \centering
 \includegraphics[width=.8\linewidth,trim=58mm 58mm 58mm 58mm,clip]{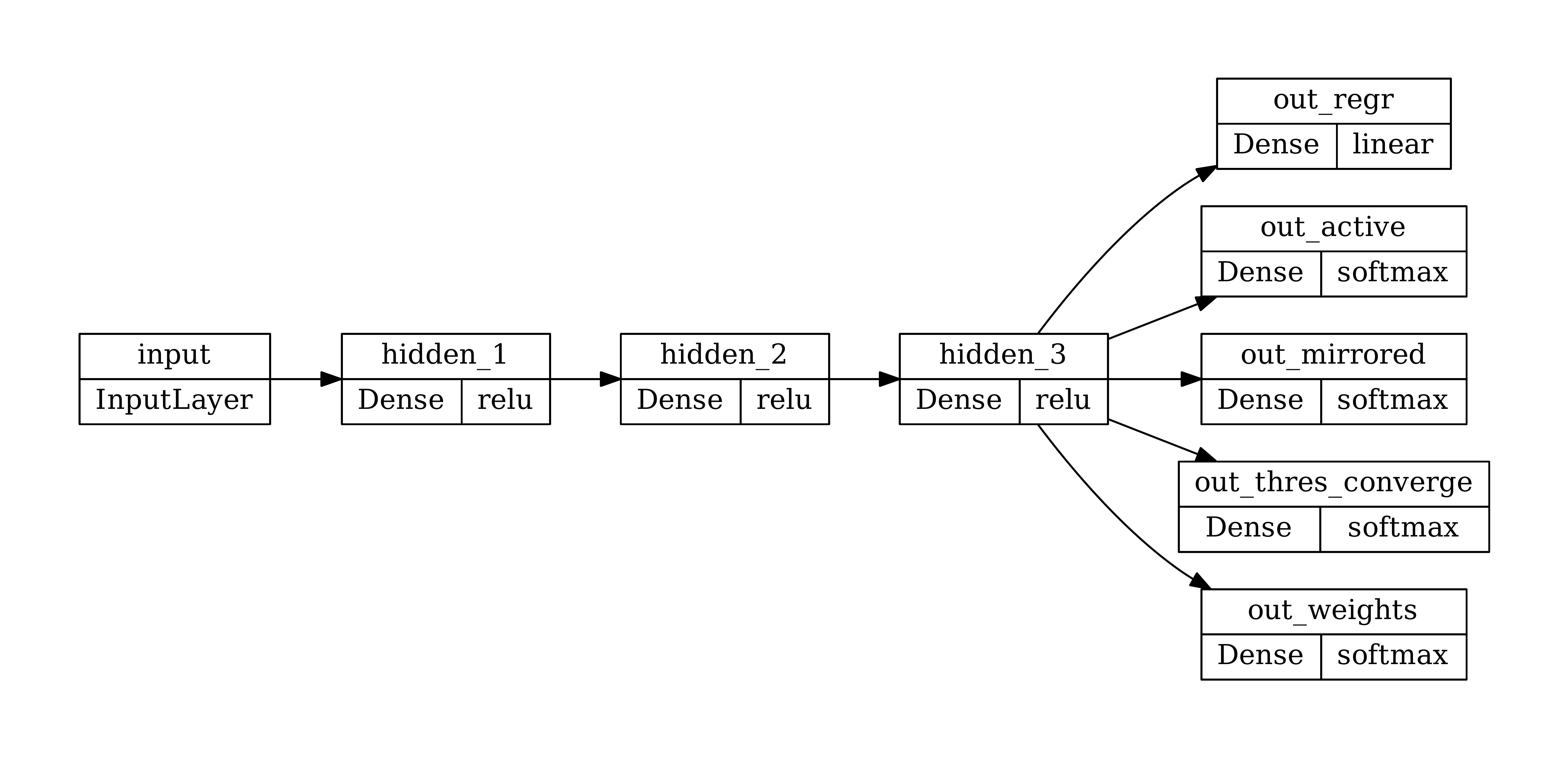}
 \caption{
 An example of the architecture of a dense NN model.
 From left to right, an input layer, three hidden layers, and several output layers, with one output layer for regression and four layers for classification tasks.
 }
 \label{fig:dnn_structure}
\end{figure}

\paragraph{Dense neural network:}
In this work, we investigate the potential of dense NN models with the following architecture for the multi-output mixed regression and classification tasks in landscape-aware AAC, as visualized in Figure~\ref{fig:dnn_structure}.

\begin{itemize}
\item \textbf{Input layer:}
The size of the input layer is equal to the number of ELA features available in the training data.

\item \textbf{Hidden layers:}
To determine an optimal inner architecture, different combinations of number of hidden layer \{$1,2,3$\}, hidden layer sizes \{$16,32,64,128$\}, and epochs \{$100,150,200$\} are evaluated using a grid search approach, $80:20$ train-test split of the training data, and a repetition of five times.
Eventually, the hidden layers are constructed based on the combination with the smallest validation loss and assigned with rectified linear unit (ReLU) as activation function.

\item \textbf{Output layers:}
In short, different layers are assigned for the mixed regression and classification tasks.
While a single output layer with linear activation function is dedicated for the multi-output regression task, the multi-output multi-class classification task is split into multiple classifications tasks.
Precisely, an output layer with softmax activation function is allocated for each categorical hyperparameter.
Consequently, the size of each output layer depends on the number of hyperparameters respectively.

\item \textbf{Loss functions:}
The dense NN models are trained using mean squared error as loss function for regression and categorical cross entropy for classification task.

\end{itemize}

\paragraph{Random forest:}
For a fair evaluation, the performance of trained NN models is compared against RF models, which are popular in landscape-aware ASP.
Precisely, the RF models are optimally constructed with fine-tuned configurations using~\texttt{auto-sklearn}~\cite{feurer2020auto}, an automated CASH tool designed for ML, and $80:20$ train-test split of the training data.
Since multi-output multi-class classification is currently limited in~\texttt{auto-sklearn}, the algorithm configuration problem is defined as a multi-target regression task, where the categorical hyperparameters are encoded as numerical labels.

\section{Experimental Setup}\label{sec:experiment}

In brief, the scope of our investigations can be summarized as follows:

\begin{itemize}
\item In $5d$, using a set of $1\,000$ RGF as training data, while the $24$ BBOB functions of the first instance as unseen test problems.
For a comprehensive analysis, we also investigate models trained using $1\,000$ MA-BBOB functions and a combination of both RGF and MA-BBOB functions;

\item  An optimization landscape is characterized based on a total of $68$ ELA features that can be computed without requiring additional function evaluations, using a DoE of $50 \cdot d$ samples, \texttt{pflacco}~\cite{prager2022pflacco}, and a similar workflow proposed in~\cite{long2024generating};

\item In this work, we consider fine-tuning the configuration of modular CMA-ES within the configuration search space in Table~\ref{tab:hp_cmaes}, with all optimization runs are allocated with a budget of $1\,000 \cdot d$ evaluations and $10$ repetitions; and

\item The TPE available in~\texttt{HyperOpt}~\cite{bergstra2013making} is employed to identify optimal configurations of modular CMA-ES and assigned with a budget of $500$ evaluations.

\end{itemize}

\begin{table*}[htbp]
  \centering
  \scriptsize
  \caption{
  An overview of the $11$ hyperparameters of modular CMA-ES considered for AAC. 
  The default configuration is highlighted in bold, where the default learning rates are automatically determined based on other hyperparameters.
  The `number of children' predicted by predictive models is rounded-off to integer.
  Symbol: $\bbbz$ for integer, $\bbbr$ for continuous variable, and $\textbf{C}$ for categorical variable.
  }
    \begin{tabular}{clcl}
    \toprule
    Num.  & Hyperparameter & Type  & Domain \\
    \midrule
    1     & Number of children & $\bbbz$ & \{ 5, …, 50 \} ($\mathbf{4+\left \lfloor (3 \cdot ln(\boldsymbol{d})) \right \rfloor}$) \\
    2     & Number of parent & $\bbbr$ & [ 0.3, 0.5 ] (\textbf{0.5}) \\
          & (as ratio of children) &       &  \\
    3     & Initial standard deviation & $\bbbr$ & [ 0.1, 0.5 ] (\textbf{0.2}) \\
    4     & Learning rate step size control & $\bbbr$ & [ 0.0, 1.0 ] \\
    5     & Learning rate covariance & $\bbbr$ & [ 0.0, 1.0 ] \\
          & matrix adaptation &       &  \\
    6     & Learning rate rank-{$\mu$} update & $\bbbr$ & [ 0.0, 0.35 ] \\
    7     & Learning rate rank-one update & $\bbbr$ & [ 0.0, 0.35 ] \\
    8     & Active update & \textbf{C} & \{ True, \textbf{False} \} \\
    9     & Mirrored sampling & \textbf{C} & \{ \textbf{none}, `mirrored', `mirrored pairwise' \} \\
    10    & Threshold convergence & \textbf{C} & \{ True, \textbf{False} \} \\
    11    & Recombination weights & \textbf{C} & \{ `\textbf{default}', `equal', `1/2$^\wedge$lambda' \} \\
    \bottomrule
    \end{tabular}%
  \label{tab:hp_cmaes}%
\end{table*}%

To analyze the performance of our approach for BBO problems in higher dimensionality, our investigations are extended to $20d$ using a smaller experimental scope to minimize computational effort, namely a DoE of $20 \cdot d$ samples for ELA features computation, $100 \cdot d$ evaluations for optimization runs, $300$ evaluations for TPE, and only the seven real-valued hyperparameters of modular CMA-ES are considered.

\subsection{Optimization Performance Metric}\label{sec:metric}

For real-world applications, (i) it is often practical to find good solutions within a shorter time, rather than finding the global optimum, and (ii) the global optimum is usually not known, making it difficult to use some popular performance metrics, e.g., expected hitting time~\cite{vermetten2020integrated}.
Hence, we propose to measure the performance of a configuration based on its area under the curve (AUC) of optimization convergence (Figure~\ref{fig:bad_rgf}).
By minimizing the AUC metric, we are essentially searching for configurations that have an optimal trade-off between the solution found and convergence speed.
In this work, all AUC during HPO are computed using the min-max normalized objective values based on the global optimum and worst DoE sample.

\subsection{Optimization Baseline}\label{sec:baseline}

Principally, we consider the following three algorithm configurations as comparison reference to evaluate the potential of our approach.

\begin{itemize}

\item \textbf{Default configuration:}
The readily available configuration in its original implementation that is simply utilized by practitioners with limited experience in fine-tuning configurations.
Inline with our motivation, our approach is primarily compared against it.

\item \textbf{SBS:}
The configuration that can perform well on average across all 24 BBOB functions and serves as our secondary target to beat in this work.
Precisely, it is identified based on the mean performance of configurations evaluated across all BBOB functions.

\item \textbf{Virtual best solver (VBS):}
The best performing configuration for a particular BBOB function, which can be treated as the lower bound.

\end{itemize}
Unlike typical ASP approaches, where the SBS and VBS are selected from a portfolio of limited algorithms using grid search, evaluating all possible configurations within the large search space in Table~\ref{tab:hp_cmaes} is computationally infeasible.
Subsequently, we determine both solvers via HPO using TPE within an allocated budget.
Due to the stochastic nature of TPE, there might be configurations that can outperform the VBS identified, but are not discovered during HPO.

\section{Results}\label{sec:result}

Due to the limited space, experimental results and figures not included in this paper can be found in our repository at~\url{https://doi.org/10.5281/zenodo.10965507}.

\subsection{Representativeness of Training Data}\label{sec:result_repre}

\begin{figure*}[!htbp]
 \centering
 \includegraphics[width=.4\linewidth,trim=0mm 0mm 32mm 0mm,clip]{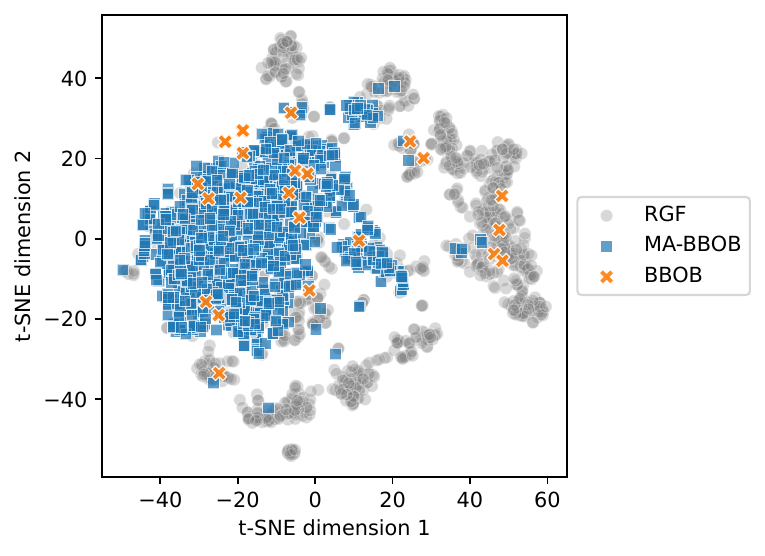}
 \hspace{3mm}
 \includegraphics[width=.53\linewidth,trim=0mm 0mm 0mm 0mm,clip]{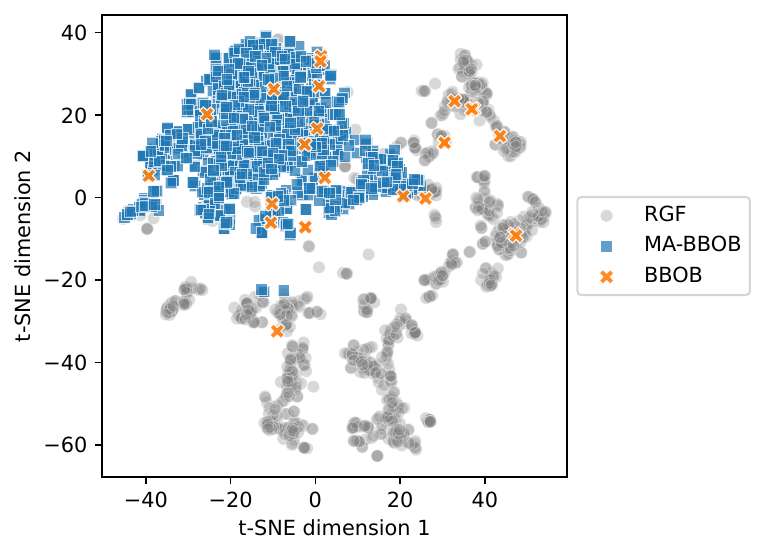}
 \caption{
 Projection of the ELA feature space to a $2d$ visualization using t-distributed stochastic neighbor embedding (t-SNE)~\cite{tsne_maaten2008} for $1\,000$ RGF, $1\,000$ MA-BBOB, and $24$ BBOB functions in $5d$ (\textit{left}) and $20d$ (\textit{right}), using a similar approach as in~\cite{long2024generating}.
 }
 \label{fig:ela_5d}
\end{figure*}

Before delving into analyzing the configuration performances, we take a closer look at the representativeness of training data.
Naturally, predictive models trained using MA-BBOB functions are expected to perform well, since the problem classes available in the training data should sufficiently cover the BBOB suite.
While this can be observed most of the time, it is not always the case, notably for F7 (step ellipsoidal) and F12 (Bent Cigar) in Section~\ref{sec:result_perform}.
The poor performances could be due to the insufficient coverage of ELA feature space by MA-BBOB functions, as shown in Figure~\ref{fig:ela_5d}, which might be related to the generation of MA-BBOB functions~\cite{vermetten2023mabbob}.
In comparison, RGF can better cover the ELA feature space, highlighting the benefits of using RGF as training data.
In fact, it seems to be advantageous to combine the large distribution of RGF and the more focused distribution of MA-BBOB on some of the BBOB functions.

\subsection{Performance of Predicted Configurations}\label{sec:result_perform}

\begin{figure*}[!htbp]
 \centering
 \includegraphics[width=.16\linewidth,trim=2mm 2mm 39mm 2mm,clip]{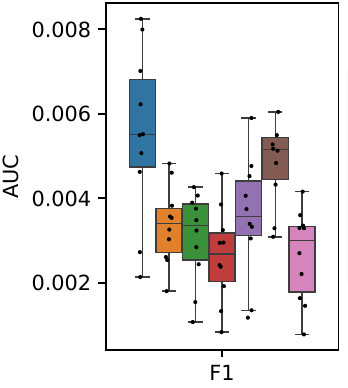}
 \includegraphics[width=.15\linewidth,trim=6mm 2mm 39mm 2mm,clip]{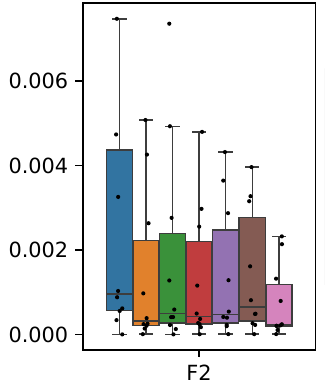}
 \includegraphics[width=.14\linewidth,trim=6mm 2mm 39mm 2mm,clip]{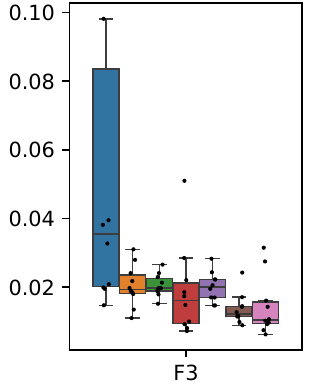}
 \includegraphics[width=.155\linewidth,trim=6mm 2mm 39mm 2mm,clip]{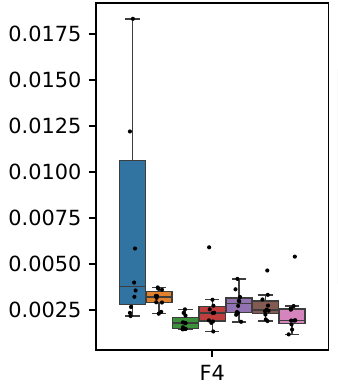}
 \includegraphics[width=.155\linewidth,trim=6mm 2mm 39mm 2mm,clip]{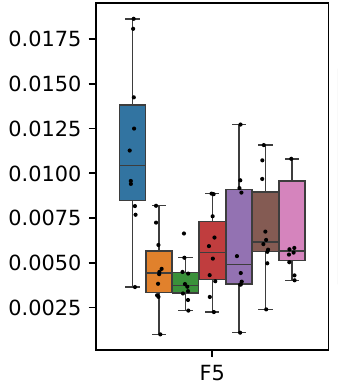}
 \includegraphics[width=.15\linewidth,trim=6mm 2mm 39mm 2mm,clip]{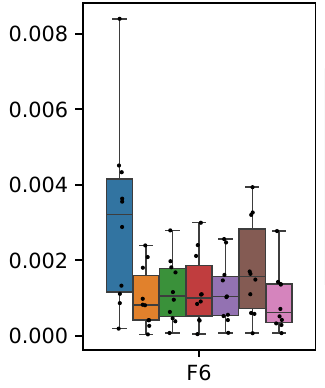}
 \includegraphics[width=.16\linewidth,trim=2mm 2mm 39mm 2mm,clip]{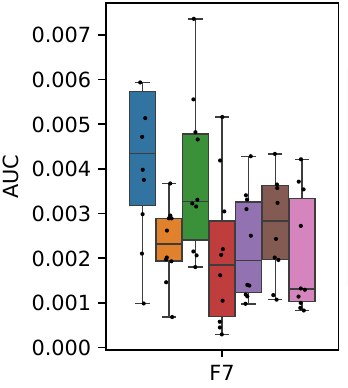}
 \includegraphics[width=.15\linewidth,trim=6mm 2mm 39mm 2mm,clip]{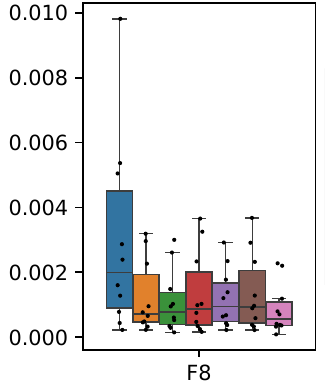}
 \includegraphics[width=.15\linewidth,trim=6mm 2mm 39mm 2mm,clip]{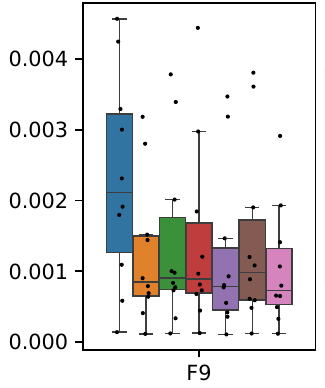}
 \includegraphics[width=.15\linewidth,trim=6mm 2mm 39mm 2mm,clip]{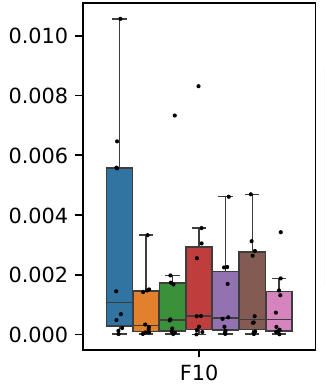}
 \includegraphics[width=.15\linewidth,trim=6mm 2mm 39mm 2mm,clip]{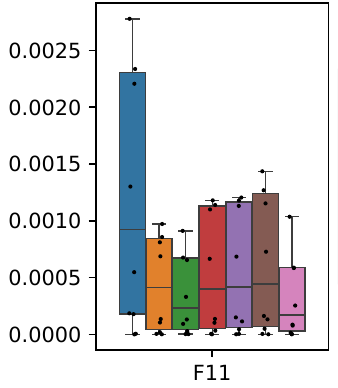}
 \includegraphics[width=.15\linewidth,trim=6mm 2mm 39mm 2mm,clip]{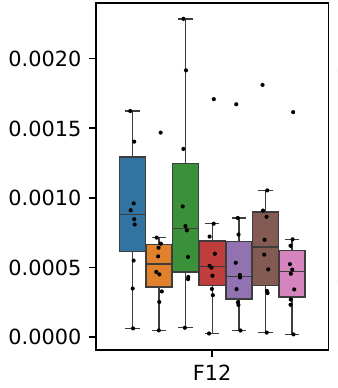}
 \includegraphics[width=.16\linewidth,trim=2mm 2mm 39mm 2mm,clip]{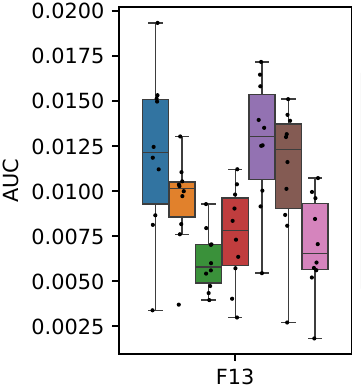}
 \includegraphics[width=.15\linewidth,trim=6mm 2mm 39mm 2mm,clip]{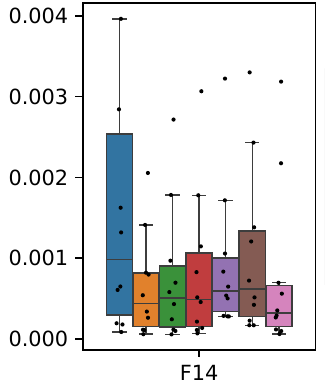}
 \includegraphics[width=.15\linewidth,trim=6mm 2mm 39mm 2mm,clip]{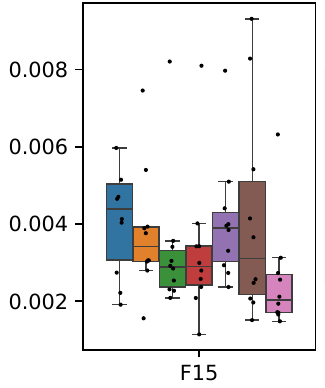}
 \includegraphics[width=.15\linewidth,trim=6mm 2mm 39mm 2mm,clip]{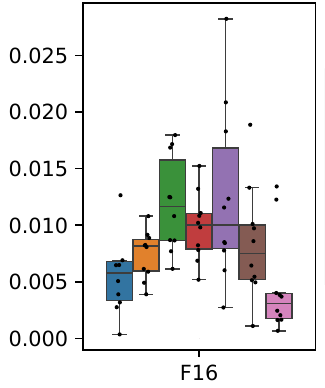}
 \includegraphics[width=.15\linewidth,trim=6mm 2mm 39mm 2mm,clip]{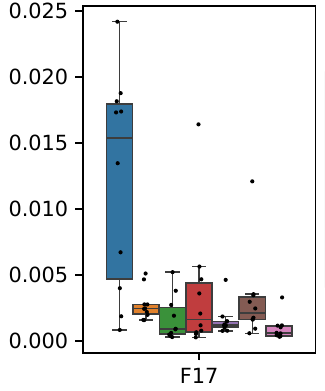}
 \includegraphics[width=.15\linewidth,trim=6mm 2mm 39mm 2mm,clip]{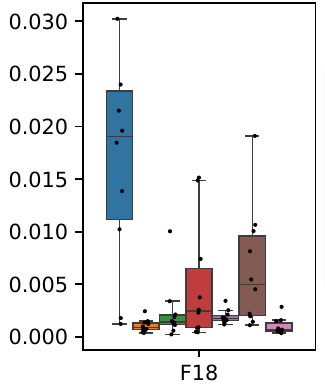}
 \includegraphics[width=.16\linewidth,trim=2mm 2mm 39mm 2mm,clip]{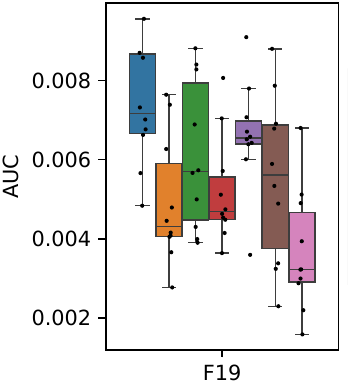}
 \includegraphics[width=.15\linewidth,trim=6mm 2mm 39mm 2mm,clip]{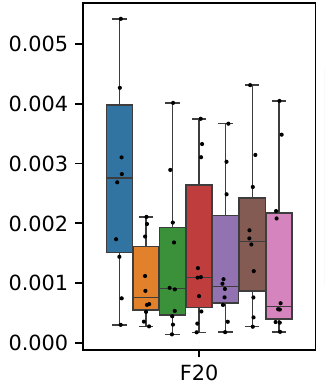}
 \includegraphics[width=.15\linewidth,trim=6mm 2mm 39mm 2mm,clip]{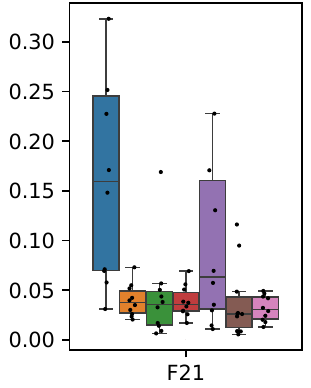}
 \includegraphics[width=.14\linewidth,trim=6mm 2mm 39mm 2mm,clip]{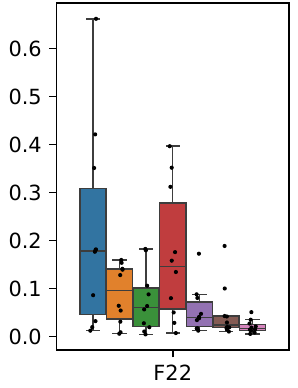}
 \includegraphics[width=.15\linewidth,trim=6mm 2mm 39mm 2mm,clip]{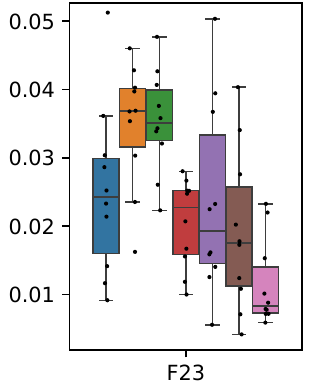}
 \includegraphics[width=.15\linewidth,trim=6mm 2mm 39mm 2mm,clip]{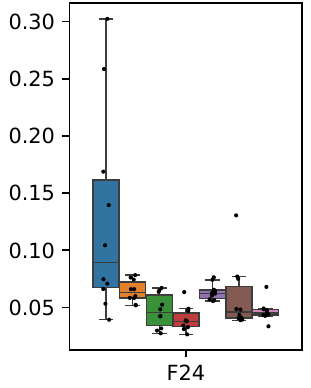}
 \includegraphics[width=1.\linewidth,trim=60mm 39mm 2mm 20mm,clip]{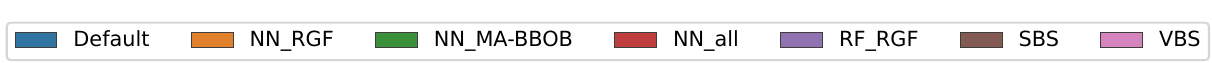}
 \caption{
 Performance of modular CMA-ES using different configurations for $24$ BBOB functions in $5d$, each repeated for $10$ times.
 The AUC is computed based on objective values min-max normalized using the global optimum and worst solution in all configurations, divided by the evaluation budget.
 A lower AUC is better.
 }
 \label{fig:ml_hpo_5d}
\end{figure*}

The optimization performances using different configurations for $24$ BBOB functions in $5d$ are compared in Figure~\ref{fig:ml_hpo_5d}.
In general, the optimal configurations identified using predictive models can clearly outperform the default configuration on most BBOB functions.
On the other hand, the predicted configurations seem to be competitive against the SBS, such as for F7 and F17 (Schaffers F7).
Not only that, our approach using NN models can perform better than the SBS in some cases, for instance, for F5 (linear slope) and F13 (sharp ridge).
Nonetheless, the performance of predicted configurations is lacking for highly multi-modal functions, e.g., F16 (Weierstrass) and F23 (Katsuura), which might be due to the absence of ELA features that can accurately capture the landscape characteristics of such complex functions, revealing the weaknesses in our approach.
When compared against the VBS, the predicted configurations sometimes seem to have a comparable performance, e.g., for F21 (Gallagher’s Gaussian 101-me peaks).

\begin{figure}[!htbp]
 \centering
 \includegraphics[width=1.\linewidth,trim=0mm 6mm 0mm 0mm,clip]{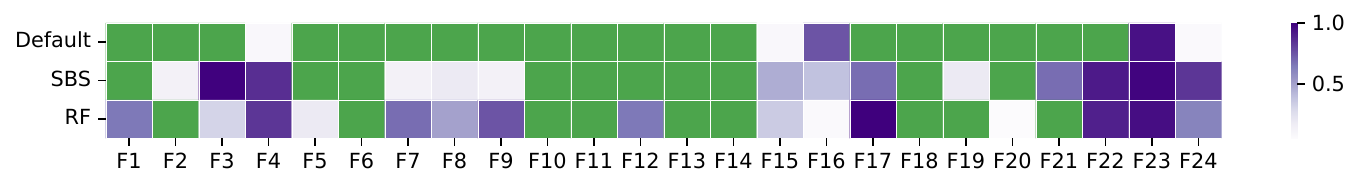}
 \includegraphics[width=1.\linewidth,trim=0mm 3mm 0mm 0mm,clip]{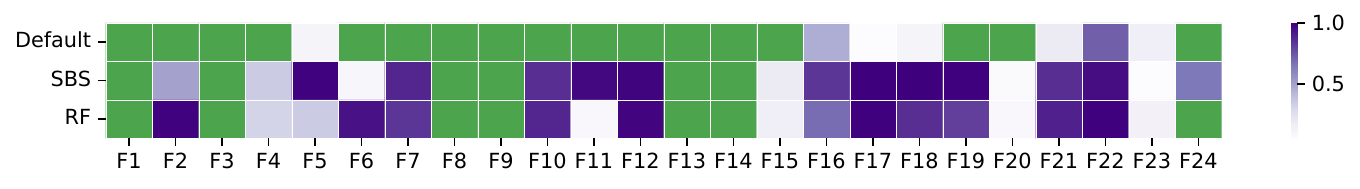}
 \caption{
 Pairwise performance comparison between the configuration predicted using NN models (our approach) against the default configuration, SBS, and RF models for $24$ BBOB functions in $5d$ (\textit{top}) and $20d$ (\textit{bottom}) based on the p-value computed using the Wilcoxon signed-rank test.
 The green color indicates that there is statistically significant evidence to support the hypothesis~\emph{optimal configurations predicted using NN models can perform better}, with a p-value smaller than $0.05$.
 Alternatively, a darker purple color (larger p-value) indicates that the hypothesis is more likely to be rejected, while a lighter purple color (smaller p-value) for a lower chance of rejection.
 }
 \label{fig:ml_hpo_stats}
\end{figure}

Using the Wilcoxon signed-rank test with the hypothesis~\emph{optimal configurations identified using our approach can perform better}, we statistically evaluate the performance of different configurations.
Precisely, we focus on comparing NN models against the default configuration, SBS, and RF models, using RGF as training data.
Inline with our previous observations, optimal configurations predicted using our approach can indeed beat the default configuration for most BBOB functions, while outperforming the SBS on many BBOB functions, as depicted in Figure~\ref{fig:ml_hpo_stats}.
It is worth reminding that our approach can be competitive against the default configuration and SBS in a few remaining BBOB functions, as previously discussed in Figure~\ref{fig:ml_hpo_5d}.
This analysis also indicates that our current approach is more effective on simple functions (first half of the BBOB suite) compared to complex functions (second half), which might be related to the ELA features.
Apart from that, the performances of NN models are as good as or even better than RF models for some BBOB functions, particularly in $5d$.

As illustrated in Figure~\ref{fig:ml_hpo_stats}, we can in general have similar observations for the BBOB functions in $20d$.
When compared to the default configuration, our approach are more effective for many BBOB functions.
Nevertheless, the performance improvements gained using our approach compared to the SBS in $20d$ are less than in $5d$, showing rooms for improvement in high dimensionality.

\section{Conclusions and Future Work}\label{sec:conclusion}

Aiming to assist practitioners unfamiliar with fine-tuning of algorithm configurations, we propose to construct general purpose predictive models towards landscape-aware AAC that can identify optimal algorithms as well as hyperparameters for different practical applications. 
To improve the generalization of our approach, we consider tree-based RGF as training data, which covers a diverse set optimization problem classes.
Furthermore, a pre-selection step is implemented to select RGF that are appropriate for AAC purposes, and thus, to improve the prediction accuracy.
Moreover, we investigate the potential of dense NN models for the multi-output mixed regression and classification tasks, which can easily handle the mixed-integer search space and large training data sets.

When evaluated on the BBOB suite in $5d$ and $20d$ using modular CMA-ES, our results reveal that we can predict near-optimal configurations that outperform the default configuration and compete against the SBS in most cases.
This is particularly encouraging for real-world applications, where such a SBS is usually not available.
In fact, properly selected RGF have promising potential as training data for landscape-aware AAC, since they cover a broader spectrum of function complexity compared to BBOB and MA-BBOB functions.
Subsequently, we believe that our approach can generalize well beyond the BBOB suite, provided that the unseen problems is well represented by the RGF training set.
Overall, configurations with better performance can be best identified using dense NN models trained on a combination of RGF and MA-BBOB functions.

For future work, we plan to improve our investigations as follows:

\begin{itemize}

\item The configuration search space can be expanded to include a variety of optimization algorithms and hyperparameters;

\item The performance of NN models can be further improved by fine-tuning more hyperparameters using an optimizer, e.g., learning rate and batch size;

\item An analysis can be extended to better understand the impact of ELA features pre-processing, e.g., using normalization vs. standardization;

\item To further minimize the overall computational costs, alternatives that can efficiently identify RGF appropriate for AAC purposes can be explored;

\item Despite the fact that the estimated $y_{opt}$ seems to be robust in our work, i.e., always smaller than all solutions found, further investigations are needed for confirmation and/or improvements; and

\item Eventually, we aim to evaluate and quantify the benefits of our approach for real-world expensive BBO problems.

\end{itemize}

\begin{credits}
\subsubsection{\ackname} 
The contribution of this paper was written as part of the joint project newAIDE under the consortium leadership of BMW AG with the partners Altair Engineering GmbH, divis intelligent solutions GmbH, MSC Software GmbH, Technical University of Munich, TWT GmbH. The project is supported by the Federal Ministry for Economic Affairs and Climate Action (BMWK) on the basis of a decision by the German Bundestag.

\subsubsection{\discintname}
The authors have no competing interests to declare that are relevant to the content of this article.
\end{credits}

\bibliographystyle{splncs04}
\bibliography{main}
\end{document}